\documentclass{ecai}
\usepackage{graphicx}
\usepackage{latexsym}

\usepackage{times}
\usepackage[dvipsnames]{xcolor}
\usepackage{latexsym}
\usepackage{subfigure}
\usepackage{amsmath}
\usepackage{amssymb}
\usepackage{adjustbox}
\usepackage{booktabs}
\usepackage{makecell}
\usepackage{multirow}
\usepackage[colorlinks=True,linkcolor=black,filecolor=black,urlcolor=black,citecolor=black]{hyperref}
\usepackage[linesnumbered,ruled,vlined]{algorithm2e}

\begin{document}

\begin{frontmatter}

\title{Towards Abstractive Timeline Summarisation using Preference-based Reinforcement Learning} 
%Preference-based Reinforcement Learning towards Abstractive Timeline Summarisation} % ES: I find that this suggestion does not scan, so I suggest an alternative above
%ES: I like this but wonder whether "preference-aligned" sounds too personalised? Brainstorming some alternatives...
%"Improving Abstractive Timeline Summaries by Aligning with Human Feedback"
%"Improving Abstractive Timeline Summaries using Preference-based Reinforcement Learning"

% UNCOMMENT FOR FINAL VERSION -- submission must be double-blind
\author[A]{\fnms{Yuxuan}~\snm{Ye}\orcid{0000-0003-1677-5196}}
\author[A]{\fnms{Edwin}~\snm{Simpson}\orcid{0000-0002-6447-1552}\thanks{Corresponding Author. Email: edwin.simpson@bristol.ac.uk.}}
% % use of \orcid{} is optional

\address[A]{Intelligent Systems Laboratory, University of Bristol}

\begin{abstract}
This paper introduces a novel pipeline for summarising timelines of events
reported by multiple news sources. 
Transformer-based models for abstractive summarisation generate coherent and concise 
summaries of long documents 
but can fail to outperform established extractive methods on specialised tasks such as timeline summarisation (TLS).
% for which they have not been trained.   
%which have received less attention from researchers. 
%For instance, timeline summarisation (TLS) is a field where old-fashioned extractive summarisation methods are still the mainstream and large pretrained summarisers haven't promoted (or shined?) yet.
While extractive summaries are more faithful to their sources, 
they may be less readable and contain redundant or unnecessary information. 
% SOLUTION OVERVIEW 
%This paper proposes a method for aligning abstractive timeline summaries with human preferences, allowing the adaptation of pretrained abstractive summarisers to TLS.
This paper proposes a preference-based reinforcement learning (PBRL) method for 
adapting pretrained abstractive summarisers to TLS,
which can overcome the drawbacks of extractive timeline summaries. 
We define a compound reward function that learns from 
keywords of interest and pairwise preference labels, % provided by users.
which we use to fine-tune a pretrained abstractive summariser via offline reinforcement learning. 
We carry out both automated and human evaluation on three datasets, 
finding that our method outperforms a comparable extractive TLS method on two of the three benchmark datasets,
and participants prefer our method's summaries 
to those of both the extractive TLS method 
and the pretrained abstractive model. 
The method does not require expensive reference summaries 
and needs only a small number of preferences to align the generated  summaries with human preferences. Code available at \url{https://github.com/Haruhi07/PBRL-TLS}.
% OLD ABSTRACT FROM ACL SUBMISSION
% TASK SETTING & MOTIVATION: 
% In recent years, abstractive summarisation methods have advanced significantly, thanks to the transformer architecture and large training datasets. 
% PROBLEM: 
%However, many specialised summarisation tasks, such as timeline summarisation (TLS), still suffer from a shortage of training data, as reference summaries are costly to create. 
% SOLUTION OVERVIEW
% This paper proposes an interactive learning method to overcome the lack of reference summaries for fine-tuning specialised summarisation models for tasks such as TLS. 
% SOLUTION DETAILS
%We define a compound reward function for evaluating generated summaries that adapts automatically to preferences provided by the user. The reward is used to fine-tune an abstractive summarisation model via offline reinforcement learning. In every interaction round, the reward function can be utilised to guide hundreds of training episodes. 
% BENEFIT OF PROPOSAL 
% Therefore, the proposed method does not rely on reference data and only requires comparative preferences and keywords of interest from an individual user, which are typically cheaper to provide. 
% EXPERIMENTAL FINDINGS
%The experiments show that our interactive training method enables the summariser to generate consistent, coherent, and fluent output for TLS tasks. Participants in 
\end{abstract}

\end{frontmatter}

\section{Introduction}
Keeping up with the news on a topic of interest involves %requires, entails
reading multiple articles from various news providers published across a range of dates. 
Timeline summarisation (TLS) makes % it easier to follow complex events 
this task easier % ; streamline this process
by presenting a chronological list of the main events related to a specific subject, distilled from multiple sources \cite{chieu2004query, tran-etal-2015-timeline}. 
Table \ref{table:examples} shows examples of timelines generated by extractive and abstractive summarisation.
% Useful alternative words; unfolding, development, progress, narrative; affair, proceedings, acts; description
In typical news summarisation tasks, 
abstractive summarisation can produce summaries of comparable quality to human authors \cite{zhang-etal-2020-pegasus}. %highly fluent, coherent and concise summaries. ES: removed 'multi-document'
However, successful pretrained models such as BART \cite{lewis2020bart} and PEGASUS \cite{zhang-etal-2020-pegasus} cannot be directly applied to TLS as,
unlike the tasks these models were trained on,
TLS requires the system to identify key dates and structure the summary into  corresponding events. %, and summarise hundreds of articles.

For specialised summarisation tasks, such as TLS, training data is in short supply, as example summaries are expensive to acquire. % time-consuming to write. %, which further limits the adoption of abstractive methods.
Therefore, existing TLS systems \cite{martschat2018temporally, 
ghalandari2020examining, li2021timeline} often rely on extractive methods %\cite{mihalcea-tarau-2004-textrank, gholipour-ghalandari-2017-revisiting2} 
\cite{lin2011class, gholipour-ghalandari-2017-revisiting2} that select sentences from the source articles and avoid fine-tuning large neural networks. 
While extractive summaries are highly faithful to their source articles, 
 concatenating pre-existing sentences may result in summaries that contain unwanted information, repeat earlier points, omit relevant information or lack fluency.
%A previous abstractive TLS method \cite{steen2019abstractive}  avoided large-scale training 
%by using word-graphs, but did 
While pretrained transformers are a potential solution, 
they have not been evaluated in previous 
abstractive TLS systems \cite{steen2019abstractive, chen2019learning,chen2023follow}.
%have not leveraged pretrained transformer models to generate fluent text. 
We therefore investigate a new approach to adapt pretrained transformers to TLS that avoids the cost of writing reference summaries. 

%<motivate the use of PL>
%If writing a good summary is time-consuming, judging the quality of a summary may be easier.
In this paper, we propose to learn a reward function 
%adapt pretrained abstractive summarisation models to TLS by learning % toooo long?
from a combination of pairwise preference labels and keywords provided by human annotators, which is used to directly optimise a pretrained summarisation model by reinforcement learning.
This combined approach is known as preference-based reinforcement learning (PBRL) \cite{cheng2011preference}. 
For a pair of timeline drafts, $s1$ and $s2$, a pairwise label $P(s1, s2)$ indicates that the annotator prefers $s1$ to $s2$. We use a set of preference labels to train a preference model  %scoring function 
that can evaluate the quality of any given summary \cite{thurstone1927law, simpson2020scalable}, and include this preference model into our reward function. 
Preference models have been shown to be highly consistent with user choices \cite{kingsley2010preference}, can outperform established metrics of summary quality that depend on reference summaries \cite{zopf-2018-estimating},
and can be learned from small numbers of preferences \cite{simpson2020scalable}.  % simpson-etal-2020-interactive}. % add this in if accepted, keep it out for now to trim reference page down!
While the preference model provides holistic guidance to the summariser,
keywords directly specify themes that are important to include in the summary. %We devise a further cOur approach 
Including keywords in the reward function therefore aims to focus the timeline when summarising a large number of news articles that cover different aspects of the same topic. 
To produce abstractive timeline summaries, we embed a pretrained abstractive summariser into a TLS pipeline, and fine-tune it using reinforcement learning with our preference-based reward, 
without the need for any reference timelines. 
%Unlike supervised learning, which encourages the model to reproduce the reference data, reinforcement learning allows the model to explore alternative generations, relying on the reward function to reward or punish these new outputs.  
Thus, learning the reward from human feedback aligns the summaries with annotator's preferences \cite{bohm2019better}, which we find results in timelines that are preferred by human evaluators over those of a closely comparable extractive method and a zero-shot abstractive summariser. 
% TO-DO: check use of the term "topical coherence". Is there an alternative phrase? There is a clash with news 'topics' in the dataset. 
% TO-DO: Is this thought too speculative? Perhaps it's one for the discussion... It's hard to learn a generation function that generalises from one reference summary to a new topic because the space of possible summaries is so huge. 
% A scoring function needs only output a single value, which could make the learning task easier. 
% TO-DO Consider whether the drawbacks of extractive methods should be mentioned earlier -- otherwise, why do we suddenly start talking about abstractive methods?

The core contributions of this work are as follows:
(1) An approach for adapting pretrained summarisation models to TLS without the need for reference timelines, 
using PBRL with a compound reward function;
(2) The first evaluation of a pretrained transformer for abstractive timeline summarisation on three benchmark datasets, finding that zero-shot performance is marginally worse than the closest extractive alternative \cite{ghalandari2020examining};
(3) We show that timelines produced after fine-tuning with PBRL have
%timelines that achieve 
higher BERTScores \cite{zhang-etal-2019-bertscore} and  
are preferred to extractive summaries by human evaluators.

\begin{table*}[htbp]
\small
\begin{center}
{\caption{Timeline examples of Libya in Arab Spring generated by extractive and abstractive systems.}\label{table:examples}}
    \begin{tabular}{l|lp{13cm}}
    \hline
    Ext. Timeline & 2011-03-02 & What's happening in Libya? Libya holds the largest crude oil reserves in Africa and oil prices rose to a two-year high today.\\
                  & 2011-03-19 & The French and British governments have lead the military intervention in Libya, despite having been among the most enthusiastic supporters of Gaddafi 's rehabilitation in the 2000s [AFP].\\
                  & 2011-04-20 & Tim Hetherington -- who was nominated for an Oscar this year with co-director Sebastian Junger for "Restrepo", a documentary about U.S. troops in Afghanistan, and Chris Hondros, a New York-based photographer for the Getty agency -- were reportedly killed in the volatile city of Misrata.\\
    \hline
    Abs. Timeline & 2011-03-02 & Crude oil rose for a second day in New York on Monday as the crisis in Libya and unrest in the Middle East continued to rattle investors.\\
                  & 2011-03-19 & France's military intervention in Libya has been described as "Sarkozy's war" by the French philosopher Bernard - Henri Levy, who helped to persuade President Nicolas Sarkozy to arm the Libyan rebels.\\
                  & 2011-04-20 & A British film director and war photographer who was nominated for an Oscar has been killed in a mortar attack in Libya.\\
    \hline
    \end{tabular}
\end{center}
\end{table*}

\section{Background}
\subsection{Event Detection}
Prior TLS systems use various methods to identify events from the article collection for a certain topic \cite{binh2013predicting, tran2015timeline, steen2019abstractive, ghalandari2020examining}. They usually work in a two-stage manner. In the first stage, the system identifies important temporal information (year, date, etc.) and assigns them to the articles. In the second stage, a summariser  generates a summary for each date. The generated summaries are combined in date order to make up a timeline. 
% ES: I don't think the statement below is useful because we (A) don't discuss or compare the alternatives to event detection and (B) have no evidence that this is how people do it.
%In this paper, we use the clustering-based event detection method since it is the most similar way that human writers summarise a timeline.

\subsection{Preference Learning from Pairwise Labels}
\label{subsec:gppl}

To optimise the summarisation model, we need an objective that reflects the quality of the summary in the eyes of the user. 
Therefore, we map pairwise labels $P$ provided by a human annotator to a score function $f$ using Gaussian Process Preference Learning (GPPL) \cite{simpson2020scalable}, an extension of the random utility model proposed by Thurstone \cite{thurstone1927law}:
\begin{equation}
    p(P(s_1,s_2)|f(s_1), f(s_2)) = \Phi(z),
\end{equation}
where $P(\cdot)$ is the pairwise preference label indicating that summary $s_1$ is preferred to $s_2$, $\Phi$ is the cumulative distribution function of the standard normal distribution which is also known as a probit likelihood, and $z=\frac{f(s_1)-f(s_2)}{\sqrt{2\sigma^2}}$. Assuming that the %posterior 
distribution of $f$ is a multivariate Gaussian, the probit $\Phi(z)$ allows us to infer the score function $f$ from pairwise labels by approximate Bayesian inference. %marginalising it. 
The Bayesian approach of GPPL permits inference with small amounts of pairwise labels, thus reducing the cost of the learning process, and accounts for contradictory and incorrect labels \cite{simpson2020scalable}.

\subsection{Reinforcement Learning: Actor-Critic}
\label{subsec:rl}
We use reinforcement learning (RL) to train the abstractive summariser since the preference-based score function is not applicable for supervised learning.
%A preference-based score function is not a suitable objective for supervised learning, so we turn to reinforcement learning (RL) to optimise the abstractive summariser.
For a given reward function $R$, at the timestep $T$, the objective $L$ can be written as the expected reward:
\begin{equation}
    L(\theta) = E_{\pi_\theta} \left[\sum_{t=0}^{T-1}R_{t+1}\right],
\end{equation}
where $\theta$ is the parameters of the policy function $\pi$. The gradient of $\theta$ can be written as follows:
\begin{equation}
    \nabla L(\theta) = \sum_{t=0}^{T-1} \nabla_{\theta}log{\pi_{\theta}(a_t|s_t)}G(t)
\end{equation}
where $G(t)=\sum_{t'=t+1}^{T}{R_{t'}}$ is the return starting from the state $s_t$, and $a_t$ is the action taken at the corresponding timestep.

Actor-Critic introduces a baseline value predicted by the critic $\hat{v}_w$ (parameterised by $w$) \cite{mnih-2016-asynchronous}.
The advantage of the action $a_t$ is computed to assess how much better it is than taking the average action at the given state $s_t$.
\begin{equation}
    Adv(t) = G(t) - \hat{v}_{w}(t)
\end{equation}
Integrating the advantage into the objective function optimises the actor's policy to yield positive returns in the long term, 
while the critic learns by minimising the mean squared error between the predicted and real values.
Therefore, the policy of the actor and the critic enhance each other through the learning process, 
and the variance of sampling from the policy distribution is less for Actor-Critic comparing to the general Policy Gradient algorithm.
Their gradients can be written as follows:
\begin{equation}
    \nabla L_{actor}(\theta) = \sum_{t=0}^{T-1} \nabla_{\theta}log{\pi_{\theta}(a_t|s_t)}Adv(t)
\end{equation}
\begin{equation}
    \nabla L_{critic}(w) = \frac{1}{T}\sum_{t=0}^{T-1}\nabla_{w}\hat{v}_{w}(s_t)
\end{equation}

\section{Our Method}
\subsection{Workflow}
\subsubsection{Baseline}

Our method follows the same workflow as CLUST \cite{ghalandari2020examining}, an extractive \emph{event detection} approach. 
%In this paper, we 
%take the state-of-the-art extractive clustering-based TLS system, CLUST \cite{ghalandari2020examining}, as our baseline. It
CLUST first encodes source documents into vectors using TF-IDF then clusters them. Each cluster is assigned the date that is mentioned most frequently by the source documents in its article collection. Then the clusters are ranked by the number of times their assigned dates are mentioned throughout the entire set of source documents,
and the top-$l$ clusters are selected as the key events. An extractive summariser, CentroidOpt \cite{gholipour-ghalandari-2017-revisiting2}, is used by CLUST to select sentences from each key event cluster and concatenate them as a summary. 

\subsubsection{Contextualised Embedding-based Event Detection}
We keep the two-stage event detection approach and change some components to better capture event information.
%in our proposed method. 
In order to group similar source documents, we use Sentence-BERT \cite{reimers-2019-sentence-bert} to compute the contextualised embedding of the source documents instead of TF-IDF,
as Sentence-BERT embeddings better reflect semantic similarity. We take the average embedding over the sentences as the representation of each document.
Prior TLS methods clustered articles using Affinity Propagation (AP) \cite{steen2019abstractive} and Markov Clustering (MC) \cite{ghalandari2020examining}.
%Affinity Propagation (AP) and Markov Clustering (MC) are the clustering algorithms used in prior TLS methods \cite{steen2019abstractive} and \cite{ghalandari2020examining} respectively. 
Therefore, we test these two algorithms as well as Agglomerative Clustering (AC) \cite{murtagh2012algorithms},
%with the contextualised embedding, 
as these algorithms do not require pre-setting the number of clusters. The date assignment and cluster ranking mechanism are kept as the same in the baseline (CLUST).

\subsubsection{Learning to summarise using PBRL}
\begin{figure}[htbp]
    \centering
    \includegraphics[scale=0.3]{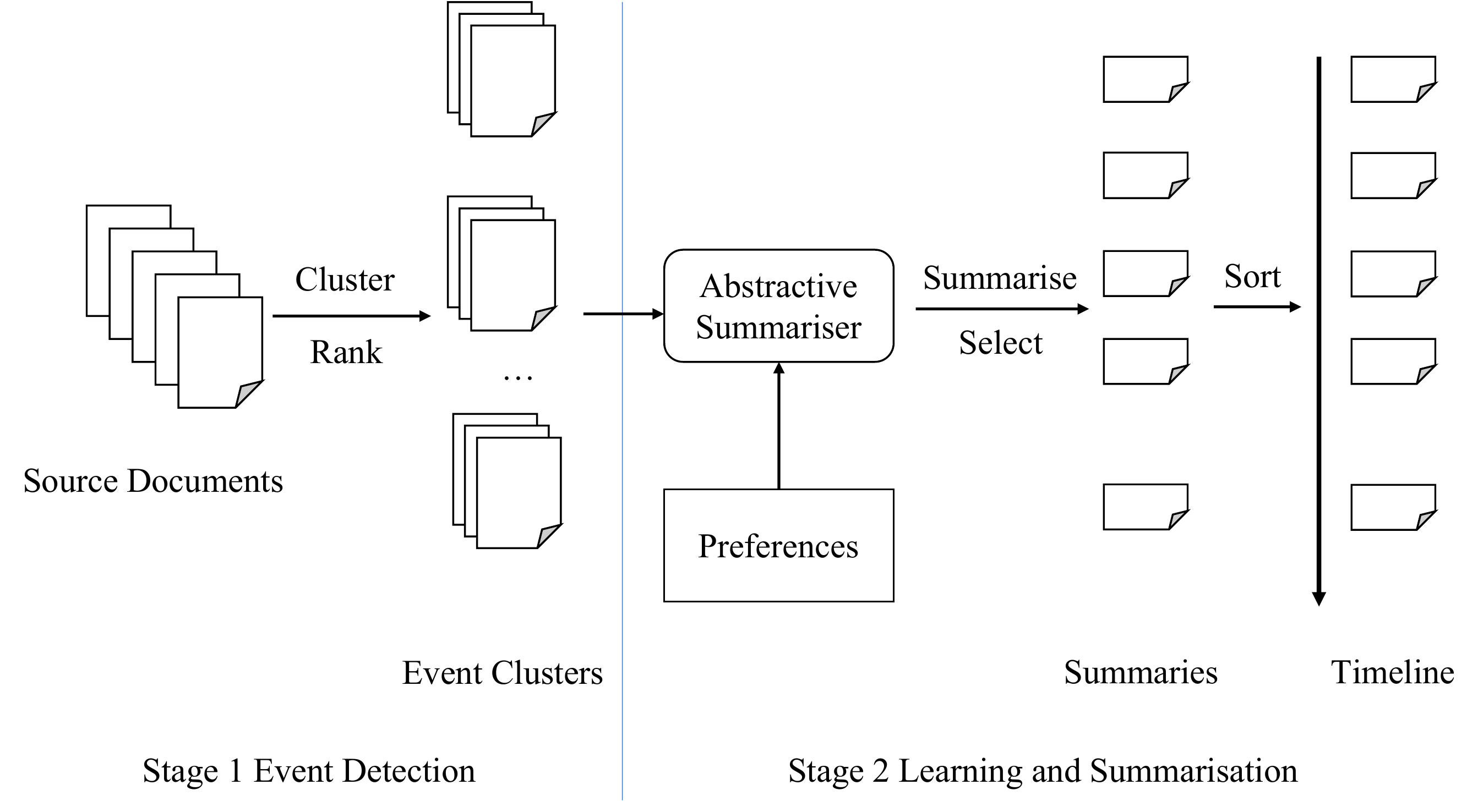}
    \caption{The system works in a two-stage manner. The first stage detects events using a clustering algorithm. The abstractive summariser learns from the preferences and summarises each event. All the generated summaries will be sorted by their date to form a timeline. }
    \label{fig:workflow}
\end{figure}

For the second stage, we integrate PBRL with an abstractive summariser implemented by a pretrained neural network.
To generate a summary for each event cluster, the summariser is applied to each cluster in turn. The source documents within each cluster are concatenated to form a single input text. 
PBRL fine-tunes the summariser leveraging the input text from each cluster. The workflow is shown in Figure \ref{fig:workflow}. The steps are as follows:
\begin{enumerate}
    \item Before the fine-tuning starts, the user reads and ranks example timelines generated by the extractive baseline and our system using the pretrained abstractive summariser in zero-shot mode. Pairwise preference labels are derived from the ranking.
    \item The user provides keywords of interest about the timeline content.
    \item The score function $f$ is learned using GPPL mentioned in Section \ref{subsec:gppl} and then used to update a sub-reward function as a component to the compound reward.
    \item The system visits each cluster in turn and samples hundreds of episodes for each individual cluster. The summariser learns from the episodes by utilising the reward function for reinforcement learning with the Actor-Critic method described in section \ref{subsec:rl}.
    \item The system generates event summries for each cluster and concatenates them in date order to form the timeline.
\end{enumerate}

\subsection{Compound Rewards}
Optimising purely for keywords and preference scores could
lead to low quality text, as the policy may find shortcuts such as repeating keywords; a score function $f$ learned from few preference labels is insufficient to guard against this. 
Therefore, some prior RL-based abstractive summarisers \cite{paulus2018a, yadav2021reinforcement}  
combined a supervised learning loss 
%computed using reference summaries 
with a reinforcement learning loss converted from a ROUGE-oriented \cite{lin2004rouge} reward. 
%to compute the reward . Their loss functions are constituted by a supervised learning loss and a 
Both losses rely on reference summaries, which may be unavailable and are
%sometimes they are not available or 
expensive to create.
This motivates us to define several sub-reward functions 
that ensure complementary qualities of the summary, including 
adherence to user preferences and interests, 
consistency with sources, and 
fluent, non-repetitive text.
%Including this combination of sub-rewards avoids the system from finding degenerate solutions, such as maximising predicted user preference score
%by generating text that is unrelated to the sources.
% ES: got to be specific here %contributing to different aspects of the summaries.

\subsubsection{Preference Reward}
To improve %obtain information for improving 
the pretrained summariser, %To obtain more information, 
our system acquires %enables users to provide 
two types of preference input: pairwise labels $P$, which indicate users' high-level preferences over different summary versions, and keywords $K$, which specify the details they prefer to see in the generated summaries. The coherence between the generated summary $s$ and the user's preferences $P, K$ is assessed through the preference reward $R_1$, which is a weighted sum of two terms:
\begin{equation}
    R_1 = w \sum_{k_i \in K} {\cos{\langle \Vec{k_i},  \Vec{s} \rangle}} + (1-w) f(s), %, P)
\label{eq:r2}
\end{equation}
where ${\cos{\langle \Vec{k_i},  \Vec{s} \rangle}}$ is cosine similarity between the embedding of timeline summary $s$ and keyword $k_i$, 
and $w$ %is a hyperparameter that 
weights keyword versus pairwise preferences.
Both embeddings are computed using sentence-BERT. For the representation of the timeline, we compute the sentence embedding for each event summary, then take the mean. 
%To compute embeddings, we use sentence- embeddings in the same way as in the event detection phase, and take the average embedding over the sentences as the representation of the timeline. 
$f(s)$ is learned using GPPL \cite{simpson2020scalable} from pairwise preferences $P$ with the average sentence embeddings as the inputs.
Learning the score function $f$ only requires about 10 %pieces of 
pairwise labels and %\cite{simpson2020interactive} and 
keywords are easy to provide,
%obtaining keywords does not require much effort,
hence the cost of preference learning is much lower than that of obtaining reference timelines through crowdsourcing. 

\subsubsection{Consistency Reward}
As well as satisfying user preferences, 
the generated summary should have high semantic similarity to the source documents to ensure it conveys similar information.  
%To capture the semantic information, 
We again compute Sentence-BERT embeddings of source articles and event summaries %(all-MiniLM-L6-v2) %ES this detail is stated in the experimental setup
\cite{reimers-2019-sentence-bert} and take the average
vectors to represent the input documents and timeline. % summaries.
%in the semantic space. 
Prior work \cite{mesgar2021improving} has successfully used cosine similarity between embeddings to quantify %as a reliable metric for quantifying 
the consistency of generated text. Thus, we use the same approach to define the consistency reward $R_2$: % as follows:
\begin{equation}
    R_2 = \cos{\langle \Vec{s},  \Vec{D} \rangle}, 
\end{equation}
where $\Vec{D}$ is the embedding of the source documents. 

\subsubsection{Language Quality Reward}
Here, we do not have reference summaries to use as teacher forcing input for the decoder, unlike prior work \cite{paulus2018a}.
Without this input, the model may degenerate to outputting topically relevant but non-fluent text. 
Therefore, we define a language quality reward to maintain the linguistic fluency of the generated output. We take the generated output $s$ as the input of GPT2 \cite{radford2019language} and use its loss to evaluate the language quality of the output. The normalisation hyperparameter $\alpha$ is the maximum loss pre-computed on the validation dataset.
\begin{equation}
    R_3 = \frac{\alpha - L_{gpt2}(s)}{\alpha}
\end{equation}

\subsubsection{Repetition Penalty}
In natural language generation, reinforcement learning agents sometimes trick for high rewards if they learned that generating specific tokens receives high rewards \cite{mesgar2021improving}. Therefore, we define $R_4$ to penalise repetitive generation.
\begin{equation}
    R_4 = 1 - \frac{\text{\#repeated\_tokens}}{\text{\#tokens}}
    \label{eq:rep_penalty}
\end{equation}

\subsection{Training}
The training process takes a small amount of preferences as input,
% In the reward learning phase, pairwise labels over timelines are mapped to preferences over the constituent event summaries for each date.
then reward function $R_1$ is learned using GPPL over the average sentence embeddings for each timeline, and used as a component of the final reward $R$ when fine-tuning the summariser.
The final reward function is computed as the weighted sum of the four sub-rewards. The weights $\gamma_{1,2,3,4}$ assigned to each sub-reward are chosen on the validation dataset.
\begin{equation}
    R = \gamma_1 R_1 + \gamma_2 R_2 + \gamma_3 R_3 + \gamma_4 R_4
\end{equation}
The final reward $R$ is applied in the Actor-Critic algorithm mentioned in section \ref{subsec:rl}, in which the actor (summariser) samples hundreds of trajectories and learns from them. The training process is shown in Algorithm \ref{alg:pbrl}.

\section{Experiments}
\subsection{Datasets}
\begin{table*}[htbp]
\begin{center}
{\caption{Statistics of three datasets, cited from \cite{ghalandari2020examining}. All numbers are the average per timeline except the first column. Date compression ratio is the timeline length divided by the number of dates. Sentences compression ratio is the total sentences in each timeline divided by the number of source documents.  }\label{table:dataset}}
    \begin{tabular}{lcccccc}
    \toprule
    Dataset & \#Timelines & \#Dates & \#Docs & Timeline Length & Date Compression Ratio & Sent. Compression Ratio\\
    \midrule
    T17 & 19 & 124 & 508 & 36 & 0.43 & 0.0117 \\ 
    Crisis & 22 & 307 & 2310 & 29 & 0.11 & 0.0005 \\
    Entities & 47 & 600 & 959 & 23 & 0.06 & 0.0017 \\
    \bottomrule
    \end{tabular}
\end{center}
\end{table*}

We conduct the experiments on three benchmark English TLS datasets, Timeline17 (T17) \cite{tran2013leveraging}, Crisis \cite{tran2015timeline} and Entities \cite{ghalandari2020examining}. T17 and Crisis are made up of news articles published by news agencies. The documents in Entities are collected from Wikipedia and the topics are celebrities' biographies. We take $1$ timeline from each dataset as the validation set to tune the hyperparameters.
Table \ref{table:dataset} reveals some of the properties of the three datasets. %ES: moved some details to the caption, think they are more useful there when reading the table. 
Crisis and Entities have larger source document pools for each timeline compared to T17, while their reference timelines are more compressed in terms of text and time, which makes it harder to find the right dates for the summaries on the two datasets.

\subsection{Model Settings and Evaluation Metrics}
In the event detection stage, we use all-MiniLM-L6-v2 \cite{reimers-2019-sentence-bert} to compute the sentence embedding. The distance threshold for Agglomerative Clustering is set to $0.7$. 
To prevent source documents being excessively long, 
we keep the setting in the baseline, CLUST \cite{ghalandari2020examining},
that takes only the first $5$ sentences of each document. 
%as the input for each document.
We took the pretrained BART-large-xsum \cite{lewis2020bart} as our baseline summariser since it was fine-tuned on a related task (news summarisation). The critic is a linear regression model with input size of $768$. We used two AdamW optimisers (learning rates $\alpha_{actor}=2e-4, \alpha_{critic}=1e-3$, $\beta=(0.9, 0.999)$ for both) to optimise the summariser (actor) and the critic in RL. We train our model on an NVIDIA 2080Ti with batch size 1. The number of episodes is 300 for each event cluster.

In terms of preferences, we simulated keyword choices for each timeline by extracting 10 terms 
%the terms keywords for each timeline 
with the highest TF-IDF scores from reference timelines, allowing us to compare with the reference summaries in the evaluation.  
Regarding the pairwise preference labels, we produced five candidate timelines for each topic and hired annotators to rank them. Pairwise labels were then derived from this real human preference ranking.

We evaluated the event detection performance by computing the date overlap between the system timeline and the ground truth. As for the generated content, we conducted the evaluation using BERTScore \cite{zhang-etal-2019-bertscore}. It replaces the exact matches in ROUGE by contextualised embeddings, resulting in more robustness as an evaluation metric for generation tasks. However, BERTScore is not able to take the temporal information into consideration, which is greatly related to the quality of a timeline. We therefore applied Alignment-based ROUGE \cite{martschat2018temporally}, which is specially adapted for TLS, to evaluate the system output from both perspectives. Additionally, we hired human evaluators to assess the legibility, factual consistency, topical coherence, and informativeness of the preferences of the system output, which are tricky for automated metrics to evaluate.

\subsection{Results}
\begin{table}
\begin{center}
{\caption{The date selection F score for different settings. CE refers to contextualised embedding. * indicates that the improvement is significant compared to the baseline ($\alpha$=0.01).}\label{table:date}}
\begin{tabular}{lccc}
\toprule
&T17&Crisis&Entities\\
\midrule
MC+TF-IDF (baseline)    & 0.407 & 0.226 & \textbf{0.174} \\
\midrule
MC+CE & 0.273 & 0.176 & 0.160 \\
AP+CE  & 0.450 & 0.205 & 0.129 \\
AC+CE (ours)   & \textbf{0.490*} & \textbf{0.239*} & 0.154 \\
\bottomrule
\end{tabular}
\end{center}
\end{table}

\subsubsection{Event Detection Performance}
We first evaluated the system's ability to detect events, measured by its date selection performance. Table \ref{table:date} indicates that our contextualised embedding-based event detection outperforms the baseline's results on Timeline17 and Crisis but has a small decrease on Entities. 
We think this is because all the source documents for each timeline in Entities are strongly related to the same topical person. The border between two events in the embedding space is therefore not as clear as in the other two datasets, since the person's name is mentioned repeatedly 
and the contextualised document embeddings therefore intermingle, %assemble 
%in the semantic space 
leading to clustering errors.
%making it challenging for the algorithm to decide the clusters. 
In addition, the huge number of candidate dates in the source documents and 
%low date compression ratio of the 
large number of dates in the reference timelines 
make it even trickier to find the correct dates. 
%Therefore, contextualised embeddings lead to the performance drop there instead of promoting the system to identify major events more accurately.
Overall, we find that Agglomerative Clustering 
performed best in combination with contextualised embeddings, 
and therefore use AC+CE in our subsequent experiments. 
%Therefore, based on these empirical 
%best fit our system
%. Therefore, %based on these empirical results, we use Agglomertative Clustering in the following experiments.
%we choose it in our final version based on the empirical results.

%The contextualised dense sentence embedding improves the event detection results on news timeline datasets (Timeline17 and Crisis). However, the biography dataset (Entities) does not benefit from it.

\begin{algorithm}
\SetKwFunction{R}{R}
\SetKwInOut{Input}{input}
\SetKwInOut{Output}{output}
\SetKwFunction{Convert}{AvgSentenceEmbedding}
\SetKwFunction{TrainGPPL}{TrainGPPL}
\SetKwFunction{UpdateR}{UpdateR1}
\SetKwFunction{ComputeReturn}{ComputeReturn}
\SetKwFunction{ComputeAdvantage}{ComputeAdvantage}
\SetKwFunction{BackwardPropagation}{BackwardPropagation}
\SetKwFunction{ComputeActorLoss}{ComputeActorLoss}
\SetKwFunction{ComputeCriticLoss}{ComputeCriticLoss}
\SetKwFunction{SampleSummary}{SampleSummary}
\SetKwFunction{Concatenate}{Concatenate}
\SetKwFunction{Critic}{Critic}
\SetKwFunction{ComputeStates}{ComputeStates}
\SetKwFunction{GenerateSummary}{GenerateSummary}
\SetKwFunction{Concat}{SortEventSummaries}
\SetKwProg{PL}{Reward learning:}{}{}
\SetKwProg{RL}{Summariser fine-tuning:}{}{}
\SetKwProg{Generation}{Generation:}{}{}
    \PL{}{
        \Input{Timelines $\{T_t\}$ used to draw pairwise preferences, the corresponding preference labels over timelines $\left\{P^{(T)}_j\right\}$, wanted keywords $\{k_i\}$}
        \ForEach{timeline $T_t$}{
            $E_t \leftarrow$ \Convert{$T_t$}\; 
        }
        $f \leftarrow$ \TrainGPPL{$\{P^{(T)}_j\}, \{E_t\}$}\;
        $R_1 \leftarrow$ \UpdateR{$f, \{k_i\}$}\;
        \Output{Preference reward function, $R_1$}
    }
    \RL{}{
        \Input{Event clusters $\{C_i\}$, with corresponding article collection $\{A_i\}$} 
        \BlankLine 
        \ForEach{cluster $C_i$}{
            $Source_i \leftarrow$ \Concatenate{$A_i$}\; 
    	\For{$j\leftarrow 1$ \KwTo $max\_episodes$}{
                $T\leftarrow$ \SampleSummary{$Source_i$}\;
                \ForEach{token $t_i$ of the trajectory $T$}{
                    $t_{1\rightarrow j} \leftarrow$ tokens in T from 1 to $j$ \; 
                    $r_j\leftarrow$ \R{$Source_i, t_{1\rightarrow j}$}\; 
                    $s_j\leftarrow$ \ComputeStates{$t_{1\rightarrow j}$}\;
                    $\hat{v}_j\leftarrow$ \Critic{$Source_i, t_{1\rightarrow j}$}\;
                    $g_j\leftarrow$ \ComputeReturn{$s_j,t_{1\rightarrow j}$}\;
                    $Adv_j\leftarrow$ \ComputeAdvantage{$g_j, \hat{v}_j$}\;
                }
                $Loss_A\leftarrow$  \ComputeActorLoss{$T, Adv$}\;
                $Loss_C\leftarrow$ \ComputeCriticLoss{$\hat{v}, g$}\;
                \BackwardPropagation{$Loss_A, Loss_C$}\;
     	}
        }
    \Output{Fine-tuned summarisation model}
    }
    \Generation{}{
    \Input{Event clusters $\{C_i\}$, with corresponding article collection $\{A_i\}$} 
        \ForEach{cluster $C_i$}{
            $s_i \leftarrow \GenerateSummary(Source_i)$ \;
        }
        $T\leftarrow$ \Concat{$\{s_i\}$ }\;
        \Output{Return timeline $T$ generated by the trained summariser for the cluster.}
    }
\caption{\label{alg:pbrl} Learning the reward function from preferences and fine-tuning the summariser for Timeline $T$.} 
\end{algorithm}

\begin{table*}[htbp]
\begin{center}
{\caption{The BERTScore and ROUGE score of different system settings. * indicates that the improvement is significant compared to the baseline ($\alpha$=0.01).}\label{table:rouge}}
\begin{tabular}{lccccccccc}
\toprule
&\multicolumn{3}{c}{\textbf{T17}}&\multicolumn{3}{c}{\textbf{Crisis}}&\multicolumn{3}{c}{\textbf{Entities}}\\
\cmidrule(lr){2-4}\cmidrule(lr){5-7}\cmidrule(lr){8-10}
                & BERT-F& AR1-F & AR2-F & BERT-F& AR1-F & AR2-F & BERT-F& AR1-F & AR2-F \\
\midrule
CLUST (baseline)   & 0.819 & 0.082 & 0.020 & 0.821 & \textbf{0.061} & \textbf{0.013} & 0.807 & 0.051 & 0.015 \\
\midrule
Ours            & \textbf{0.822*} & \textbf{0.085*} & \textbf{0.023*} & \textbf{0.831*} & 0.059 & \textbf{0.013} & \textbf{0.821*} & \textbf{0.053*} & \textbf{0.016} \\
\quad w/o PBRL  & 0.816 & 0.082 & 0.019 & 0.823 & 0.053 & 0.012 & 0.811 & 0.047 & 0.014 \\
\quad w/o R1    & 0.820 & 0.077 & 0.018 & 0.829 & 0.056 & 0.013 & 0.817 & 0.044 & 0.011 \\
\quad w/o R2    & 0.817 & 0.073 & 0.018 & 0.830 & 0.058 & 0.011 & 0.818 & 0.045 & 0.013 \\
\quad w/o R3+R4 & 0.811 & 0.071 & 0.015 & 0.825 & 0.054 & 0.012 & 0.815 & 0.042 & 0.012 \\
AC+CentroidOpt  & 0.813 & 0.078 & 0.018 & 0.821 & 0.056 & 0.012 & 0.806 & 0.043 & 0.010 \\
\bottomrule
\end{tabular}
\end{center}
\end{table*}

\subsubsection{BERTScore}
Although ROUGE \cite{lin2004rouge} is the most commonly used evaluation metric for generation tasks, it is also challenged for being less favourable to abstractive summarisation \cite{ng2015better} as it can incorrectly penalise 
%due to possible
paraphrasing %in the system output, 
because it counts overlapping lexical units in two pieces of text. Therefore, we evaluated our system timeline's content using BERTScore, 
which uses contextualised embeddings to avoid the need for exact lexical matches.
%as a more sophisticated evaluation metric compared to ROUGE. 
As input to the metric, we concatenated the event summaries in date order for each timeline.
%as a whole to put in the metric. 
Our method outperforms the extractive baseline on BERTScore, and improves over the pretrained summariser without fine-tuning, which 
shows that abstractive approaches can outperform extractive methods when adapted to TLS.
%proves our system is capable of generating adequate timeline summaries.

\subsubsection{Adapted ROUGE for TLS}
The adapted ROUGE (marked as AR1 and AR2) matches the dates in the generated and reference timeline before computing traditional ROUGE scores. The temporal information also affects the results, because it only computes token overlaps on matched event summaries. Therefore, we use it to comprehensively evaluate the quality of our system's timelines. Results in Table \ref{table:rouge} show that our method outperforms the extractive baseline on Timeline17 and Entities, while having a slightly lower but still competitive AR1 result on Crisis. 

Since our system outperforms the extractive baseline on both BERTScore and date selection, we believe that paraphrasing in the abstractive summaries may lead to the decrease in AR1 on Crisis. Another probable reason for the lower ROUGE scores on Crisis is the huge number of documents and extremely low sentence compression ratio for each timeline. Condensing massive input into a short but adequate abstractive summary can be an extremely tricky task compared to selecting several sentences near the cluster centroid for the extractive method to cover the main idea.

\subsubsection{Human Evaluation}
We invited 10 volunteers to assess the timelines %to get a thorough comprehension of the improvement of the system 
from a comprehensive perspective. 
The participants were requested to mark timelines from 0 to 10 for two randomly-selected topics from T17 and two from Crisis. Each topic had 3 timelines generated by different TLS systems (baseline, ours, ours without PBRL),
meaning that each volunteer read and annotated approximately 80 event summaries. %with zero-shot summariser). 
Separate scores were given for %four aspects:
\emph{legibility}, 
\emph{informativeness} of the content, 
\emph{coherence} to the topic, 
and \emph{factual consistency}. 
We also encouraged the volunteers to use their own standards to give an \emph{overall} score. 

The results in Table \ref{table:human} demonstrate a
%that over the first four aspects, the 
major improvement over the extractive baseline in terms of legibility, indicating that our system generates more readable summaries.
%, which proves the positive effect of our system in generating more readable timeline summaries. 
The improvement in overall scores indicated that our system's output is preferred by the human evaluators. The student t-test between our model and the baseline demonstrates that the improvement in these two aspects is statistically significant. For these two aspects, we computed
inter-annotator agreement by mapping the scores to rankings and computing pairwise Cohen's $\kappa$. We obtained coefficients of 0.68 for legibility, 0.62 for overall score, and \textasciitilde0.53 for other all aspects, indicating moderate agreement between users. % on these two aspects.

\begin{table*}
\begin{center}
{\caption{The human evaluation score for different TLS systems. Higher score means better performance on corresponding aspect.}\label{table:human}}
\begin{tabular}{lccccc}
\toprule
               & Legibility & Informativeness & Topical Coherence & Factual Consistency & Overall\\
\midrule
Ext. Baseline  & 5.8        & 8.4             & \textbf{9.2}               & 8.3                 & 6.7\\
Ours           & \textbf{7.9}        & \textbf{8.8}             & 8.7               & \textbf{8.6}                 & \textbf{8.5}\\
\quad w/o PBRL & 6.7        & 7.7             & 8.1               & 7.2                 & 7.6\\
\bottomrule
\end{tabular}
\end{center}
\end{table*}

\subsection{Ablation Study}
To better understand the effect of each module in our system, we carried out an ablation study on PBRL and the sub-reward functions.
We evaluated the zero-shot setting on the pretrained abstractive summariser, and tested the contribution of the sub-reward functions by setting each one to zero (R3 and R4 we ablated jointly since they both aim to ensure linguistic fluency). We also tested the combination of our clustering method with the extractive summariser used in CLUST, CentroidOpt, 
to reveal whether the improvements arise purely through the changes we made to event detection.

The results in Table \ref{table:rouge} indicate that 
%simply applying the summariser without learning can lead to competitive results, 
the abstractive summariser with zero-shot setting can receive competitive results,
potentially because the summariser was previously fine-tuned on a similar task.
However, PBRL is still needed to surpass the extractive baseline. The results in the next three rows, where each sub-reward is turned off in turn, are all worse than that with the full reward function, demonstrating that 
human preference feedback is helpful in improving the summariser 
but must be balanced with topical consistency and, especially, language quality. 
%every sub-reward function is essential. 
The bottom row in Table \ref{table:rouge} shows that the extractive summariser used in CLUST performs worse than the zero-shot abstractive summariser when combined with our clustering method. Considering this result alongside Table \ref{table:date} shows that 
%which suggests that our clustering method benefits the abstractive summariser, we find that 
the clustering approach needs to be adapted to the chosen summarisation approach.

\subsection{Case Study}
We conducted a case study to further understand the characteristics of the generated summaries. We selected a topic in Timeline17 and generated three timelines using the extractive baseline (CLUST), our method with and without PBRL. They are displayed with the reference timeline in Table \ref{table:case}. There are %messy 
web formatting tokens marked in \textcolor{red}{red} in the reference timeline (\textit{-LRB-, -RRB-, \textbackslash u00ac}, etc.),
since these summaries were extracted from webpages \cite{tran2013leveraging}. 
%Moreover, the quotation marks in both reference and extractive timelines are asymmetrical -- the front quotation mark is actually double grave accent marks. 
These tokens may cause our method's performance to be underestimated,
%be partly responsible for worsening 
as it does not generate these reference tokens.
Excluding these formatting tokens makes the abstractive summaries more legible than the extractive summaries.
%At the same time, the abstractive summaries are more
%legible because not corrputed by the messy tokens as extractive timelines do.
%
The \textcolor{blue}{blue} text highlights irrelevant information %pieces to the topic 
such as page update times and the news agency name. 
The extractive TLS system fails to filter out page update information within the summaries. 
%Moreover, the quotation marks in both reference and extractive timelines are asymmetrical -- the front quotation mark is actually double grave accent marks. 
The two abstractive timeline summaries are much more concise compared to the extractive one 
as they simply state the event and avoid over-quoting details. 
%. Every summary simply states the event instead of over-quoting details in the source documents. 
%Additionally, they are much legible because not corrputed by the messy tokens as extractive timelines do. 
%
However, the abstractive summariser can generate hallucinations (marked in \textcolor{ForestGreen}{green}). The false fact is corrected in the PBRL version, demonstrating that our method may help with the factual consistency of the output, which is concurrent with similar findings for learning from human feedback \cite{ouyang2022training}.

\begin{table*}
\footnotesize
\begin{center}
{\caption{Timeline examples generated by different systems on the involvement of Conrad Murray in the death of Michael Jackson. }\label{table:case}}
    \begin{tabular}{p{1.5cm}|lp{13cm}}
    \hline
    Reference & 2009-06-25 & Dr Murray finds Jackson unconscious in the bedroom of his Los Angeles mansion. Paramedics are called to the house while Dr Murray is performing CPR, according to a recording of the 911 emergency call. He travels with the singer in an ambulance to UCLA medical center where Jackson later dies. \\
    & 2009-07-22 & The doctor's clinic in Houston is raided by officers from the Drug Enforcement Agency \textcolor{red}{-LRB-} DEA \textcolor{red}{-RRB-} looking for evidence of manslaughter. \\
    & 2010-02-08 & Dr Murray is charged with involuntary manslaughter. He pleads not guilty and is released on \$75,000 \textcolor{red}{-LRB- \textbackslash u00ac \#~} 48,000 \textcolor{red}{-RRB-} bail. The judge says he can continue to practice medicine, but bans him from administering anesthetic agents, ``specifically propofol''. \\
    & 2011-01-04 & Preliminary hearings begin. Prosecutors allege that Dr Murray ``hid drugs'' before calling paramedics on the day Jackson died. They also state that he did not perform CPR properly and omitted to tell paramedics that he had given Jackson propofol. \\
    & 2011-11-03 & The case against Dr Murray goes to the jury following closing statements. The prosecution concludes by saying the doctor's care of Jackson had been ``bizarre''. The defense maintains Dr Murray was not responsible and that the singer caused his own death while his doctor was out of the room. ``If it was anybody else , would this doctor be here today?'' defense lawyer Ed Chernoff says. \\
    & 2011-11-07 & Dr Conrad Murray is found guilty of involuntary manslaughter after nine hours of jury deliberations. The doctor was remanded in custody without bail until he receives his sentence. \\
    & 2011-11-29 & Dr Conrad Murray is sentenced to four years in county jail. Judge Michael Pastor says the evidence in the case showed him guilty of a ``continuous pattern of lies and deceit''. \\
    \hline
    Extractive Baseline & 2009-06-25 & Dr Conrad Murray is on trial accused of the involuntary manslaughter of singer Michael Jackson. \textcolor{blue}{29 September 2011 Last updated at 15:44 GMT} Help Live coverage of the trial of Michael Jackson's personal physician, Dr Conrad Murray, who is charged with involuntary manslaughter of the singer.\\
    & 2009-07-22 & Michael Flanagan of the DEA describes the operation Police have searched the Las Vegas home and offices of Michael Jackson's doctor as part of a manslaughter investigation into the singer's death. Dr Conrad Murray, who police say is not a suspect, was at Jackson's mansion and tried to revive him before he died.\\
    & 2010-02-08 & Dr Murray has denied he caused Michael Jackson's death Los Angeles prosecutors will file a criminal case against Michael Jackson's doctor on Monday in connection with the singer's death, officials say.\\
    & 2011-01-06 & Prosecutors say his negligence led to Jackson's death aged 50 in June 2009. Michael Jackson doctor Conrad Murray \textcolor{red}{`}hid drug dose' Dr Murray, shown here in February, joined Jackson's entourage to get him fit for London concerts Michael Jackson's doctor tried to hide the fact that he had given the singer a powerful anesthetic shortly before he died, a prosecutor has said.\\
    & 2011-11-03 & Conrad Murray decides not to give evidence Defense lawyers for Dr Conrad Murray say they will call no further witnesses Dr Conrad Murray has told a judge he will not testify in his own defense in the trial over Michael Jackson's death. Dr Murray told trial Judge Michael Pastor he had made his decision \textcolor{red}{``}freely and explicitly'', reports said. \\
    & 2011-11-07 & \textcolor{blue}{Page last updated at 08:14 GMT, Tuesday, 8 November 2011} Propofol is a fast-acting hospital sedative used before anaesthetics Michael Jackson's physician, Doctor Conrad Murray, has been found guilty of involuntary manslaughter by a jury in Los Angeles. \textcolor{blue}{8 November 2011 Last updated at 06:28} Michael Jackson's doctor, Conrad Murray, found guilty Michael Jackson 's doctor has been found guilty of causing the singer 's death.\\
    & 2011-11-29 & The trial of Michael Jackson's doctor, Conrad Murray, took place on the ninth floor of the superior court of Los Angeles. \textcolor{blue}{The Atlantic} Murray will be sentenced on 29 November, and was held without bail after his conviction The last doctor of Michael Jackson's life was convicted of manslaughter by a California jury because the law demanded it and the facts supported it.\\
    \hline
    Ours w/o PBRL & 2009-06-25 & The personal doctor of Michael Jackson has pleaded not guilty to involuntary manslaughter in connection with the singer's death.\\
    & 2009-07-22 & The US Drug Enforcement Administration (DEA) has searched the Los Angeles clinic of Michael Jackson's doctor.\\
    & 2010-02-08 & Dr Conrad Murray, the man accused of killing Michael Jackson, has returned to work at a Las Vegas hospital.\\
    & 2011-01-06 & Michael Jackson's doctor Conrad Murray ordered a bodyguard to pick up vials of medicine before phoning for help on the day he died, his trial has heard.\\
    & 2011-11-03 & The jury in the involuntary manslaughter trial of a doctor accused of giving Michael Jackson a fatal dose of propofol has retired to consider its verdict.\\
    & 2011-11-07 & The trial of Michael Jackson's doctor took place on the ninth floor of a Los Angeles court.\\
    & 2011-11-29 & The judge in the case of Michael Jackson's former doctor Conrad Murray has said \textcolor{ForestGreen}{he will not sentence him to prison under a new California law}.\\
    \hline
    Ours & 2009-06-25 & The personal doctor of Michael Jackson has been ordered to stand trial for involuntary manslaughter over the singer's death of a drug overdose. \\
    & 2009-07-22 & The Drug Enforcement Agency (DEA) has obtained a warrant to search the home of Michael Jackson's doctor, Conrad Murray.\\
    & 2010-02-08 & The former doctor of Michael Jackson has pleaded not guilty to the singer's manslaughter.\\
    & 2011-01-06 & Michael Jackson's doctor ordered a bodyguard to pick up vials of medicine before phoning for help, a court in Los Angeles has heard.\\
    & 2011-11-03 & The jury in the trial of a Los Angeles doctor accused of involuntary manslaughter over Michael Jackson's death has been sent home for the weekend.\\
    & 2011-11-07 & The trial of Michael Jackson's doctor has been taking place in Los Angeles.\\
    & 2011-11-29 & Michael Jackson's former doctor Conrad Murray has been sentenced to four years in prison for involuntary manslaughter.\\
    \hline
    \end{tabular}
\end{center}
\end{table*}

\section{Related Work}
%What do I want to conduct here? The proposed method is able to transfer a relatively large summariser to the specific TLS topic while aligning with human-preferable reading habits
\paragraph{Timeline Summarisation} Prior mainstream TLS methods are usually extractive and avoid %without the involvement of 
neural abstractive summarisation because the amount of the data is insufficient for training \cite{chieu2004query, ghalandari2020examining, li2021timeline}. 
Some work built up relatively large datasets to train the neural timeline summariser in a supervised manner \cite{chen2019learning, chen2023follow}, but the data is not openly accessible and each timeline is sourced from a single article in an encyclopedia, rather than multiple news sources. %I think "open access" is the key word
%not directly accessible. 
%Hence, the dates are more coarse-grained (e.g., just years) and easier to identify. 
While the evaluation used the standard ROUGE that does not consider the date structure of the timeline, the results obtained by Chen et al.\cite{chen2019learning} show that abstractive summarisation may be able to outperform extractive methods for TLS when adapted correctly to the data.
Other work on abstractive TLS managed to obviate model training by utilising the word-graph \cite{steen2019abstractive}. 
Since some prior abstractive TLS methods worked on different datasets or applied their own adapted evaluation metrics, it was not possible to directly compare our results against theirs. 
To the best of our knowledge, this paper is the first to apply a pretrained summarisation model to TLS
and to consider PBRL as a way to align timeline summaries with human preferences, instead of relying on large amounts of training data.  
%in TLS  and doesn't require large amount of training data. In addition, it reveals that PBRL can improve the quality of system timeline summaries while aligning with human-preferable reading habits.

\paragraph{Preference-based Reinforcement Learning} 
While early work on PBRL showed pairwise preferences provide more consistent rewards for RL than absolute scores \cite{christiano_NIPS2017} and that small amounts of human feedback are sufficient to enable offline RL \cite{christiano_NIPS2017, kreutzer2018reliability}.
PBRL has been 
used to refine single-document summarisation \cite{bohm2019better}, multi-document summarisation \cite{gao2020preference} and machine translation \cite{kreutzer2018reliability}, 
but these approaches did not use a compound reward function to incorporate keywords and maintain language quality. For conversational agents, a combined reward was shown to be necessary to satisfy competing objectives such as fluency and topicality \cite{mesgar2021improving},
and has been used to ensure that large language models satisfy user preferences \cite{ouyang2022training}. In summary, 
the related work does not 
adopt PBRL to TLS.
%and has not made use of compound rewards alongside preference learning. 

\section{Conclusion}
In this paper, we proposed a novel TLS pipeline. 
We used contextualised embeddings and applied a more suitable clustering algorithm to better capture the semantic information in the article collection, thus enhancing the system's ability to detect events in news datasets. We leveraged PBRL to fine-tune the pretrained abstractive summariser, improving its performance comparing to the zero-shot setting while minimising the need for training data. Our system outperforms the state-of-the-art clustering-based extractive TLS baseline in terms of BERTScore across all three datasets and achieves better ROUGE scores on two datasets. The human evaluation indicated that our abstractive timelines are more legible than those of the extractive baseline, and that overall, they align more closely with human preferences than summaries produced by the zero-shot abstractive summariser and the extractive baseline.

% ADD BACK IN FOR CAMERA READY
%\ack High performance computing -- BluePebble.
%CSC scholarship. 

% \section{Page limit}
% %The traditional page limit for ECAI long papers is {\bf 7 (six)} pages
% %in the required format. The traditional page limit for short
% %submissions is {\bf 2} pages.
% %
% %However, these page limits may change from one ECAI to
% %another. Consult the most recent Call For Papers (CFP) for the most
% %up-to-date page limits.

% The page limit for ECAI scientific papers is {\bf 7} pages, plus one ({\bf 1})
% additional page for references only. Scientific papers should report on substantial novel results. The reference list may start earlier than page 8, but only references are allowed on this additional eighth page.
% The page limit for ECAI highlights is {\bf 2} pages. They are intended for disseminating  recent technical work (published elsewhere), position, or open problems with clear and concise formulations of current challenges.

% \section{Tables}
% Tables are set in 8 point (2.8 mm) on a body of 10 point (3.5 mm).
% The table caption is set centered at the start of the table, with
% the word Table and the number in bold. The caption is set in medium
% with a 1 pica (4.2 mm) space separating it from the table number.

% A one line space separates the table from surrounding text.

\clearpage

\section*{Acknowledgements}
We thank the reviewers for their valuable comments. This work was carried out using the computational facilities of the Advanced Computing Research Centre, University of Bristol - \url{http://www.bristol.ac.uk/acrc/}. The financial support for Yuxuan Ye was provided by the programme of the China Scholarship Council (No. 202108060154).
% TO-DO The references need to fit onto one page. Let's go back and remove a few that are not essential. 
% Entries for the entire Anthology, followed by custom entries
\bibliography{anthology,custom, ecai}

\end{document}